\documentclass[runningheads]{llncs}
\usepackage{graphicx}
\usepackage{amsmath,amssymb} 

\begin{document}

\title{Learning Deeply Supervised Good Features to Match for Dense Monocular Reconstruction\thanks{Supported by the ARC Laureate Fellowship FL130100102 to IR and the Australian Centre of Excellence for Robotic Vision CE140100016.}} 
\titlerunning{Learning Deeply Supervised Good Features to Match} 


\author{Chamara Saroj Weerasekera \inst{1,2} \and
Ravi Garg \inst{1,2} \and
Yasir Latif \inst{1,2} \and Ian Reid \inst{1,2}}
%

\authorrunning{C. S. Weerasekera et al.} 


\institute{University of Adelaide, Australia 
\and
ARC Centre of Excellence for Robotic Vision
\email{firstname.lastname@adelaide.edu.au}}

\maketitle

\begin{abstract}
Visual SLAM (Simultaneous Localization and Mapping) methods typically rely on handcrafted visual features or raw RGB values for establishing correspondences between images. 
These features, while suitable for sparse mapping, often lead to ambiguous matches in texture-less regions when performing dense reconstruction due to the aperture problem. 
In this work, we explore the use of \emph{learned} features for the matching task in dense monocular reconstruction.
We propose a novel convolutional neural network (CNN) architecture along with a deeply supervised feature learning scheme for pixel-wise regression of visual descriptors from an image which are best suited for dense monocular SLAM.
In particular, our learning scheme minimizes a multi-view matching cost-volume loss with respect to the regressed features at multiple stages within the network, for explicitly learning contextual features that are suitable for dense matching between images captured by a moving monocular camera along the epipolar line. 
We integrate the learned features from our model for depth estimation inside a real-time dense monocular SLAM framework, where photometric error is replaced by our learned descriptor error. Our extensive evaluation on several challenging indoor datasets demonstrate greatly improved accuracy in dense reconstructions of the well celebrated dense SLAM systems like DTAM, without compromising their real-time performance.

\keywords{Mapping  \and Visual Learning \and 3D Reconstruction \and SLAM.}
\end{abstract}
\section{Introduction}

Visual 3D reconstruction finds its uses in many vision-based autonomous systems, including robotic navigation and interaction. It is a core component of structure from motion and visual SLAM systems that work by establishing correspondences between multi-view images, either explicitly or implicitly, and use this information to solve for the locations of both the camera and 3D landmarks. Depending on the method used, and also the application, the reconstructed 3D map varies from being sparse to dense. While sparse mapping is usually sufficient in cases where the desired output is the trajectory of a moving robot, a dense 3D map is useful when a robot is required to reliably interact and move in a given environment. Success of both dense mapping and direct dense tracking given a map crucially depends on the ability to generate accurate dense correspondences between images. Reliance on hand-crafted features or raw RGB pixel values is the major bottleneck of dense SLAM as matching ambiguities arise due lack of unique local texture, repetitive texture in the image, appearance distortion due to change in perspective, change in lightning, motion blur and occlusions.
Priors such as smoothness are often used to ``fill in'' or rectify parts of the map, but they are not always appropriate or sufficient for accurate reconstructions.

\begin{figure}[!t]
\centering
\includegraphics[width=\textwidth]{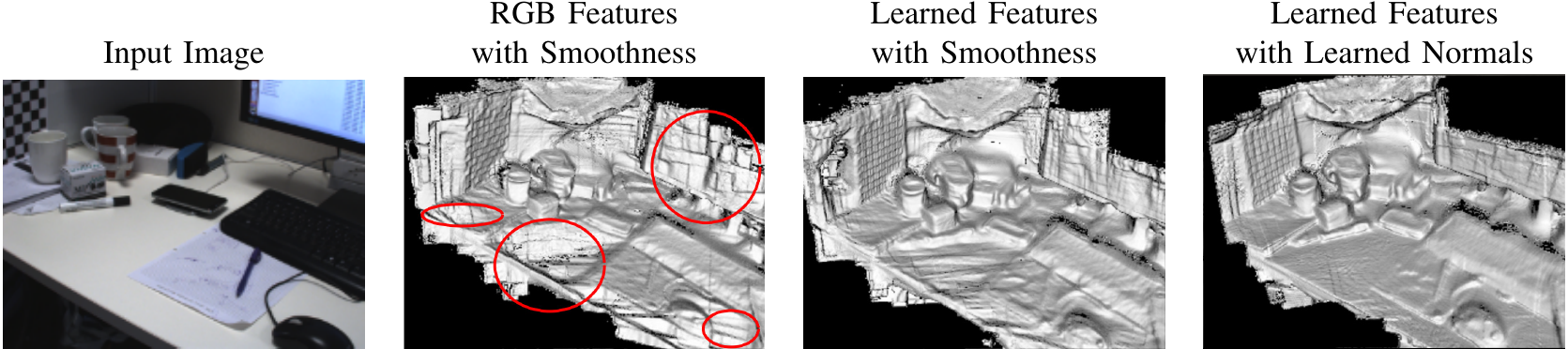}
\caption{Replacing color features with learned deep features and hand crafted priors with learned priors improves the reconstruction quality. Notice the dent in table, the very structured error in fusing the depths of the monitor and wrongly reconstructed notebook in the reconstruction which is obtained by using a RGB cost volume with smoothness regularization of DTAM \cite{newcombe2011dtam} --- all of which are fixed using our approach.}
\label{fig:intro_main}
\end{figure}

\begin{figure}[!t]
\centering
\includegraphics[width=\textwidth]{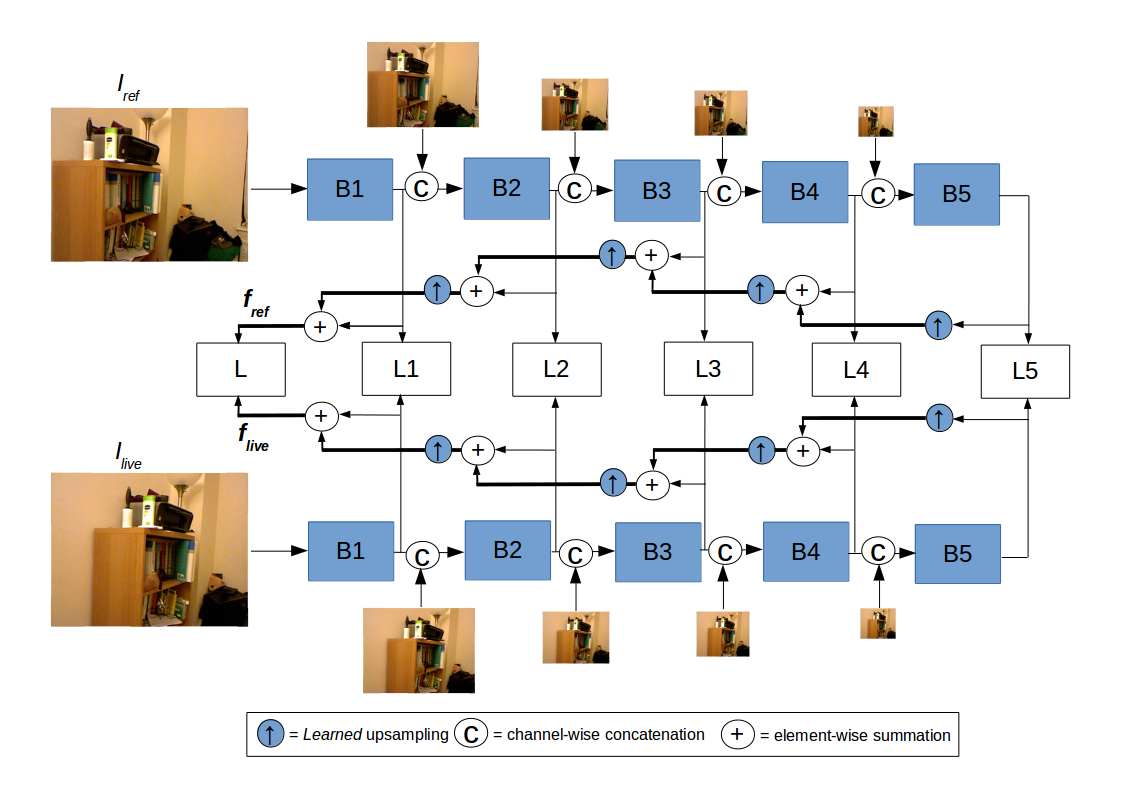}
\caption{Overview of the proposed multi-scale network architecture and training setup.
At train-time both the reference and live images ($I_{ref}$ and $I_{live}$) are passed into the network as a batch in two independent streams that share weights. The final 32 dimensional output features ($\mathbf{f_{ref}}$ and $\mathbf{f_{live}}$) for the reference and live image respectively as well as their intermediate output features are used as input to the cost volume loss layers L,L1-L5 for deep feature supervision at every image scale. Note that bold lines do not intersect with thin lines. The forward time for the network for a single image to produce a dense feature map is only $\approx 8 ms$ on a Nvidia GTX 980 GPU.\label{fig:NetArch}}
\end{figure}

Recently, the resurgence of CNNs and their capacity to capture rich scene context directly from data have allowed relatively accurate predictions of surface normals and depths to be made just from a single image \cite{eigen2015predicting,laina2016deeper,garg2016unsupervised}. CNNs have also shown to be capable of accurately solving low-level vision problems like predicting dense image correspondences \cite{OpticalFlowSpatialPyramid,KendallEndToEnd17} or depth and motion from two-views \cite{DEMON}, outperforming traditional vision methods which rely on handcrafted features and loss functions. However they have mainly been used for solving single or two-frame problems. It is not straightforward to extend standard end-to-end learning frameworks to use data from arbitrary number of views without increasing memory and computational requirements while training and testing. 

In a typical visual SLAM system where the camera captures images at 30Hz, it is always beneficial to make use of all the information available. In this work, we take a different approach of learning dense features \emph{good for matching} which are \emph{fast} to compute and can directly be used in existing dense monocular reconstruction systems for reliable correspondence estimation.

More specifically we want to automatically learn features via a CNN to deal with the ambiguities that occur when matching along epipolar lines for per-pixel depth estimation, especially when presented with images captured with a handheld monocular camera. To this end we introduce a novel deeply supervised CNN training scheme for feature descriptor learning, and a fully convolutional multi-scale network design that can help efficiently capture both local and global context around every pixel in the image to regress for the corresponding high dimensional feature vector for these pixels. In contrast to previous feature learning approaches, we propose to construct and minimize a multi-view cost volume loss for feature learning, where the predicted feature for \emph{each} pixel in the reference image is compared with those for a range of pixels along the corresponding epipolar line in the live frame (one of which will be the right match). 

The minimization is done with respect to predicted features over the \emph{entire} cost volume during training and thus millions of feature matching instances are incorporated into a single training batch which makes the training very efficient. The efficient computation of our loss is enabled by the large parallel compute capability of GPUs. We apply our cost volume loss after every feature map downsampling stage in order to regularize the network to learn good features to match at every image scale. As our training loss mimics the cost volume loss desired to be minimised by state-of-the-art dense mapping systems, our belief is that it helps the network learn the optimal features suited for dense correspondence in a monocular reconstruction setting. Similar to \cite{SchmidtVD} our training framework is self-supervised, requiring only raw RGB-D data (like that captured by the Kinect sensor) and the camera trajectory that can be estimated from the data itself using a method like RGB-D ORB-SLAM \cite{ORBSLAM2} or Kinect Fusion \cite{KinectFusion} for rigid scenes. 

We also integrate our feature descriptors into the real-time dense monocular reconstruction system in \cite{DenseSLAMNormals} which uses in addition a CNN-based learned surface normal prior for depth regularization (Fig. \ref{fig:intro_main}). We show that the combined system leads to further improvement in dense monocular reconstructions in challenging indoor environments. To the best of our knowledge this combined system is the first framework for real-time monocular dense reconstruction to harness the learning capabilities of CNNs in terms of \emph{both} geometric scene understanding (in the form of learned normals) and dense correspondence (in the form of good features to match) given arbitrary number of input images.

\section{Background}
\noindent\textbf{SLAM methods with hand crafted features:}
Most popular SLAM methods like PTAM \cite{klein2007parallel} and ORB-SLAM \cite{ORBSLAM2} use handcrafted features successfully for sparse matching and solving for a sparse map and camera trajectory. LSD-SLAM \cite{engel2014lsd} uses direct RGB alignment for solving for both pose and map, and is able to generate semi-dense maps aided by depth regularization. DTAM \cite{newcombe2011dtam} is also based on direct RGB alignment but is able to generate fully dense depth maps by accumulating multiple frames for keyframe reconstruction, and utilizing a smoothness prior to help resolve matching ambiguities. While recent work has begun exploring the use of learned priors for reconstruction \cite{DenseSLAMNormals,Facil2016,tateno2017cnn,bloesch2018codeslam}, these methods still rely on low-level color/feature matching for dense correspondence.

\noindent\textbf{End-to-end CNNs for two-frame reconstruction:}
Ummenhofer et al. \cite{DEMON} propose to train a network for depth, camera motion, and optical flow prediction from two frames for indoor scenes and
Kendel et.al. \cite{KendallEndToEnd17} propose to train a deep network end-to-end for disparity prediction outdoors, to give state-of-the-art results for stereo reconstruction in NYU and KITTI datasets respectively. These two-frame reconstruction architectures however are not very easily extensible to use more than two frames -- a limitation we address in this work. 
In particular, our work is inspired by \cite{KendallEndToEnd17}, which
constructs a cost-volume with learned features for two frames by simply concatenating the features of these images for all possible candidate matches. This stereo cost-volume is further passed to a deep 3D convolutional network which implicitly acts as a regularizer to finally select the correct matches out of these candidates. Instead of learning this regularization network, in our work we propose to make the matching decision on a per-pixel basis directly based on the cost-volume by forcing the correctly matched features to have small distances (and large otherwise). At test time we can then infer the depth of each pixel in the reference image \emph{based on the same/a similar distance metric in the learned feature space summed over multiple live frames}, which end-to-end stereo networks listed above do not allow.

\noindent\textbf{Feature learning with CNNs for dense correspondence:}
Feature learning methods designed for good matching e.g. MC-CNN \cite{LECunPatchBased} or LIFT \cite{LIFT} are often patch-based Siamese like architectures. Unlike these methods, our architecture is fully convolutional allowing for efficient dense feature extraction during training and testing, with flexibility to match each pixel relying potentially on a receptive field as large as the entire image. Also related to our work is \cite{SchmidtVD} that proposed a self-supervised visual descriptor learning method for matching. At test time, their learned features are used for model to frame alignment under extreme scene variations. Another related work is \cite{UCN16} that propose a Universal Correspondence Network for geometric and semantic feature matching. In both \cite{SchmidtVD,UCN16}, a correspondence contrastive loss is employed which minimizes the feature distance between positive matches and pushes that of negative matches to be at least margin $m$ away from each other. The cross-entropy loss that we employ enforces a similar effect by forcing the \emph{probability} of the true match to be $1$ only at the corresponding location in the other image, and $0$ elsewhere, without having to explicitly specify a margin $m$.

Some other main differences exist between our feature learning method and previous work: (1) we use deep supervision (we minimize the feature matching error at different scales in the network to learn multi-scale features that are good to match), (2) we design and use a significantly faster and efficient neural net architecture specially suited for dense correspondence (which is also independent of spatial transformer layers as used in \cite{UCN16} and \cite{LIFT16} to add to efficiency), and (3) previous work on feature learning have relied upon randomly selected negative samples (e.g. nearest neighbor sampling around the true match at \emph{integer} pixel locations in \cite{UCN16}) --- in our case, by sampling uniformly a fixed number of times along epipolar lines we generate positive and negative training examples at \emph{sub-pixel} level (through bilinear interpolation in feature space) and this is more reflective of the type of matching during dense reconstruction. Moreover, existing methods have not been extensively analyzed within/integrated into a real-time dense reconstruction framework.

\section{Method}\label{sec:method}
In this section we describe our neural network architecture and training and inference time setup, and discuss the motivations behind the design choices.

\subsection{Network Architecture}
Our network consists of five blocks B1-B5 (Fig. \ref{fig:NetArch}), each block consisting of 3 convolutional layers. All convolutional layers have a kernel size of 3x3 and 32 output filters, which we found to be a good middle ground for computational speed and matching accuracy. While we have tried other variations of our network with convolutional kernel sizes 9x9, 7x7, 5x5, the one with convolutional kernel size 3x3 in each block enabled us to achieve the highest matching precision. We believe that this is because explicitly limiting the receptive field, especially at the earlier layers, helps the network learn features that are invariant to appearance distortion.

The first two layers of each block consist of ReLU activation units for adding non-linearity to the network. The first layer of each block has a stride of 2 for 2x downsampling after each subsequent block. Downsampling in Block B1 is optional but adds to the efficiency, at a slight trade-off of precision. The progressive down sampling rapidly increases the receptive field for a pixel without the need for a deeper network and helps the network capture contextual information which is key for successful matching. Not having a very deep network also prevents the network from overfitting to a particular dataset, and helps it learn more generic descriptors that are useful for matching even in completely different types of environments (Fig. \ref{fig:kitti}). In our network, the maximum potential receptive field for a pixel is roughly half the input image resolution which is significantly larger than in work like \cite{LECunPatchBased} which limit the receptive field to 9x9 patches.

While coarse features are useful for matching large textureless regions, they are too abstract for fine-grain pixel level matching required for precise depth estimation. Inspired by the U-Net architecture \cite{UNet} we upsample each coarse prediction and sum it with the preceding fine predictions (Fig. \ref{fig:NetArch}) to produce our final feature output. Doing so explicitly embeds local information about a pixel into its contextual representation enabling precise sub-pixel level matching. 

The final output of our network is therefore a 32 dimensional feature vector for each pixel in the image, encoding both contextual and low-level image information. Each upsampling operation on the output of each coarse block is done through a single learnable deconvolution layer with a kernel size of 5x5 and stride of 2. Making this layer learnable results in improved results over a simple bilinear upsampling, showing that in this case it is better to allow the network to learn a more appropriate upsampling method. Furthermore, concatenating a bilinearly downsampled version of the input image with the output features from each block before passing them as input into the subsequent coarser blocks also improves the matching result. This shows that there is still information present in the low-level features that can complement the abstract ones in order to benefit the pixelwise matching at lower resolutions. All input RGB images (with values in the range of 0-255) are mean (of training set) subtracted but \emph{not} normalized as this gave the best performance in practice.

Although the multi-scale features from our network can be used independently for course-to-fine matching at test time, we perform matching using just the final \emph{aggregated} high resolution feature output of our network instead as the contextual information contained in it allows for a large basin of convergence and makes the solution less sensitive to initialization (Section \ref{sec:eval_all}). The fast forward time of our network ($\approx 8ms$ on a Nvidia GTX 980 GPU) makes it suitable for use as a feature extractor in real-time SLAM.
Another advantage of our architecture is that it provides the flexibility for one to discard the courser blocks if needed, as the features that are generated by each block are \emph{explicitly trained} to be suitable for matching through deep supervision (see Section \ref{training}). For example one can choose to remove Blocks B2-B5 and simply use features from Block B1 if further efficiency is desired without compromising too much on accuracy (Fig. \ref{fig:kitti}). One can also reduce the feature vector length from 32 to further improve the matching speed or increase the length to improve the matching accuracy. In our CUDA run-time implementation we make use of $float4$ data structures and 3D texture memory for improving GPU memory access times.

\subsection{Training}
\label{training}
The output of our network are dense pixelwise features for a given image. Given a pair of images (that are 30 frames apart in our training setup), the goal is to learn suitable features that give a minimum matching cost \emph{only} between a pixel $\mathbf{u}_p$ in the reference image and the corresponding matching pixel in the live image, that lies on the corresponding epipolar line and is defined by the depth of $\mathbf{u}_p$ and the relative camera motion.
To do this, first we discretize the (inverse) depth space $\rho_p$ into $K=256$ bins ($\rho_p=\{ \rho^1_p,...,\rho^K_p \}$, with $\rho_p^1=0$ and $\rho_p^K=4$), and convert the continuous matching problem into a $K$ class classification problem.

More specifically, our goal is to find the optimal features $\mathbf{f}_{ref}=f(\mathbf{w},I_{ref})$ and $\mathbf{f}_{live}=f(\mathbf{w},I_{live})$ that are a function $f$ of the shared neural network weights $\mathbf{w}$ and $I_{ref}$ and $I_{live}$ respectively, that maximize the estimated matching probability $P_{\phi}(\rho_p^l)$ (derived from the cost volume $E_{\phi}$ as defined further below) at $\rho_p^l=\rho_p^*$ (the groundtruth inverse depth label), and minimize the probability elsewhere. We achieve this by minimizing the following cross-entropy loss:
\begin {equation}
\sum_{p}\sum_l P(\rho_p^l) (- ln P_{\phi}(\rho_p^l,\mathbf{f}_{ref}, \mathbf{f}_{live})) 
+(1-P(\rho_p^l)) (- ln (1-P_{\phi}(\rho_p^l,\mathbf{f}_{ref}, \mathbf{f}_{live}))
\label{eqn:cross_entropy_loss}
\end{equation}
where, $P_{\phi}(\rho_p^l,\mathbf{f}_{ref}, \mathbf{f}_{live}) = \frac{e^{-E_{\phi}(\rho_p^l,\mathbf{f}_{ref}, \mathbf{f}_{live})}}{\sum_{k=1}^K e^{-E_{\phi}(\rho_p^k,\mathbf{f}_{ref}, \mathbf{f}_{live})}} = \sigma(-E_{\phi}(\rho_p^l,\mathbf{f}_{ref}, \mathbf{f}_{live}))$, \\ 
$E_{\phi}(\rho_p, \mathbf{f}_{ref}, \mathbf{f}_{live})=||\mathbf{f}_{ref}(\mathbf{u}_p) - \mathbf{f}_{live}(\pi(T_{nr}\pi^{-1}(\mathbf{u}_p,\rho_p)))||_2^2
$ and $\sigma(.)$ is the soft-max operation. $P(\rho_p^l)=1$ if $\rho_p^l=\rho_p^*$ and $P(\rho_p^l)=0$ otherwise.

$T_{nr}\in\mathbb{S}\mathbb{E}(3)$ is a matrix describing the transformation of a point from camera coordinates of $\mathbf{I}_r$ to that of $\mathbf{I}_n$. $\pi(.)$ is the projection operation, and $\pi^{-1}(.,.)$ is the back-projection operation, such that $\pi^{-1}(\mathbf{u}_p,\rho_p) = K^{-1}\dot{\mathbf{u}}_p/\rho_p$, where $K$ is the camera intrinsics matrix, and $\dot{\mathbf{u}}_p=(u,v,1)^T$ is $\mathbf{u}_p$ (a pixel in the reference image) in homogeneous form.

Note that since the length of the epipolar line in the live image (for the specified inverse depth range) can vary depending on the camera baseline and we sample a fixed number of bins, we perform bilinear interpolation in feature space to obtain features corresponding to each inverse depth hypothesis at non-integer pixel locations. Additionally, since the epipolar line for a pixel in the reference image can stretch out of the live image, we set a high cost $m$ (e.g. $m=10$) for negative matches that fall outside. If, on the other hand, the positive match for a pixel in the reference image falls outside of the live image, we completely mask out the loss for that pixel in the reference image.

Furthermore, in order to mitigate the effects of discretization and enforce a more refined match we minimize the regression loss in the same manner as \cite{KendallEndToEnd17}, by performing a dot product between the discrete inverse depth labels and the corresponding set of probabilities $P_{\phi}(\rho_p^l,\mathbf{f}_{ref}, \mathbf{f}_{live})$, to get a single $\hat{\rho}_p$ for each pixel, and then minimising the squared distance of $\hat{\rho}_p$ to the groundtruth inverse depth $\rho_p^*$. On top of that we also minimize the squared distance of the inverse of $\hat{\rho}_p$ ($\hat{d}_p$) to the inverse of $\rho_p^*$ ($d^*_p$). In other words we want to make the expected inverse depth/depth value to be as close as possible to the groundtruth.

Our overall loss function could therefore be written as:
\begin {equation}
\begin{aligned}
&\sum_{p} \sum_l P(\rho_p^l) (- ln P_{\phi}(\rho_p^l,\mathbf{f}_{ref}, \mathbf{f}_{live})) + (1-P(\rho_p^l)) (- ln (1-P_{\phi}(\rho_p^l,\mathbf{f}_{ref}, \mathbf{f}_{live})) \\
+ & \sum_{p} \Big( \lambda_{\rho} (\hat{\rho}_p - \rho_p^*)^2 + \lambda_d (\hat{d}_p - d_p^*)^2 \Big) 
\end{aligned}
\label{eqn:cross_entropy_loss}
\end{equation}
where, $\hat{\rho}_p = P_{\phi}(\rho_p^l,\mathbf{f}_{ref}, \mathbf{f}_{live}) \rho_p^l$. We experimentally found that using a combination of the aforementioned cross-entropy loss and the two regression losses in both depth and inverse depth space, improved the precision of $\hat{\rho}_p$. We set $\lambda_{\rho}=5,\lambda_d=1$ in order to balance the magnitude of the three types of losses, and to balance the bias in precision between depth and inverse depth label space.

\begin{figure}[!t]
\centering
\includegraphics[width=0.95\textwidth]{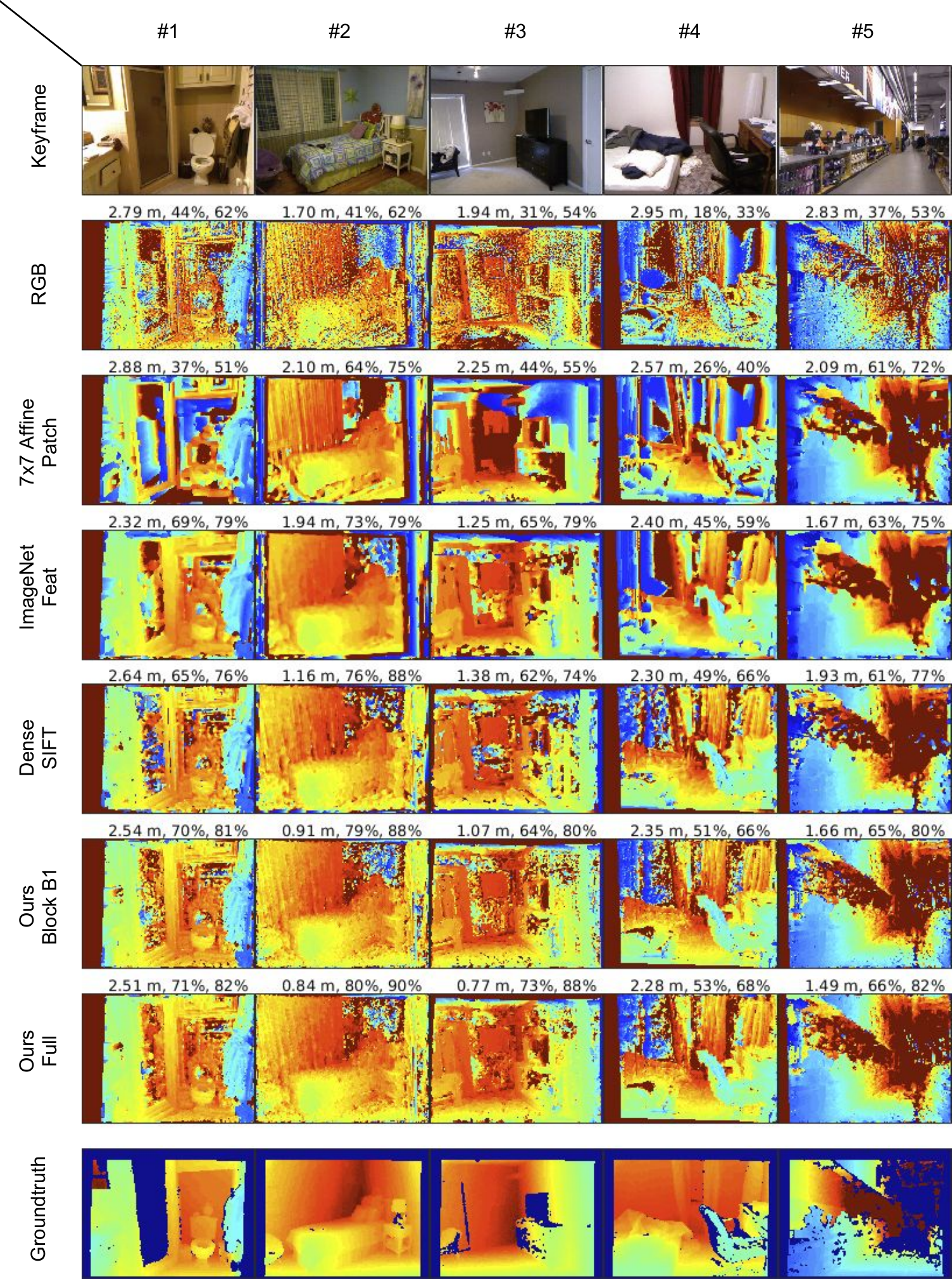}
\caption[Comparison of keyframe reconstructions (based purely on matching with 30 frames) using our learned features against those of baselines]{Comparison of keyframe reconstructions (based purely on matching with 30 frames) using our learned features against those of baselines. The three numbers above each result denote RMSE (m), high precision accuracy ($\delta < 1.1$), and medium precision accuracy ($\delta<1.25$). (Please refer supplementary material for more examples.)}
\label{fig:FeatComp1_1}
\end{figure}

\begin{figure}[!t]
\centering
\includegraphics[width=.75\textwidth, trim={2.cm 1cm 2.2cm 0},clip]{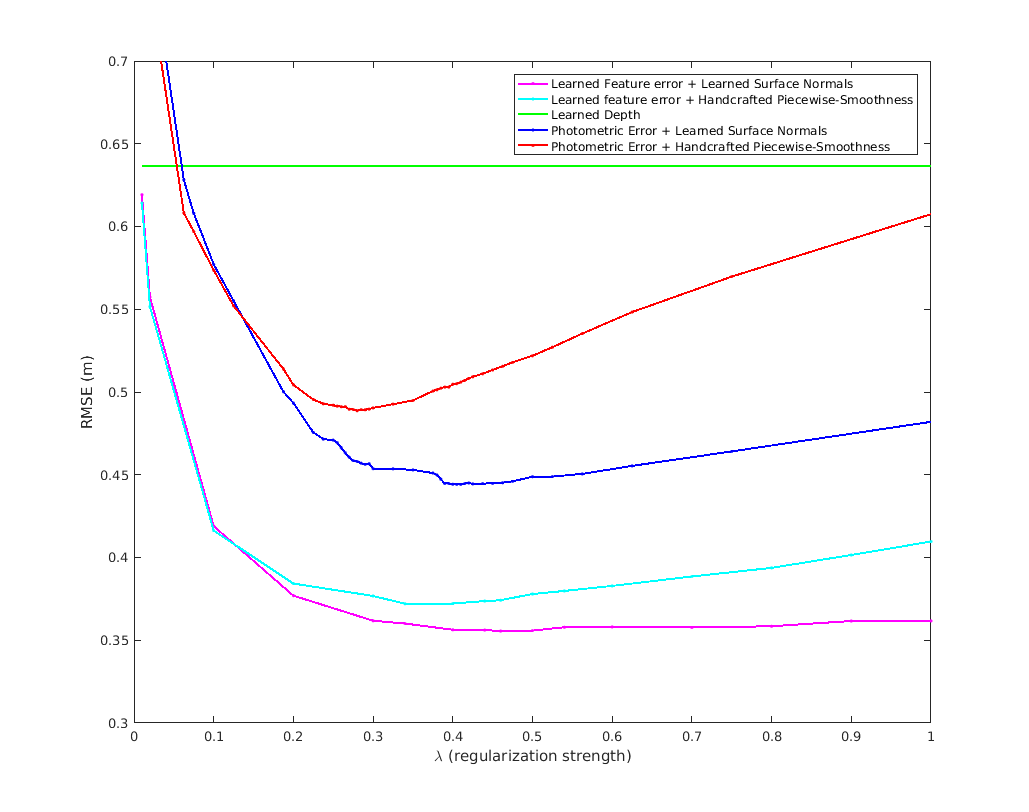}
\caption{Plot of RMSE (m) vs regularization strength ($\lambda$) for our method (bottom two curves) as well as our baselines. 
Note that as the cost-volume magnitude depends on the length and range of the used features, for the 2 plots using learned feature error the regularization strength  $\lambda$ is divided by a constant factor of 12.5 for better visualization.}
\label{fig:VaryingLambdaNYU}
\end{figure}

\begin{figure}[!t]
\centering
\includegraphics[width=\textwidth]{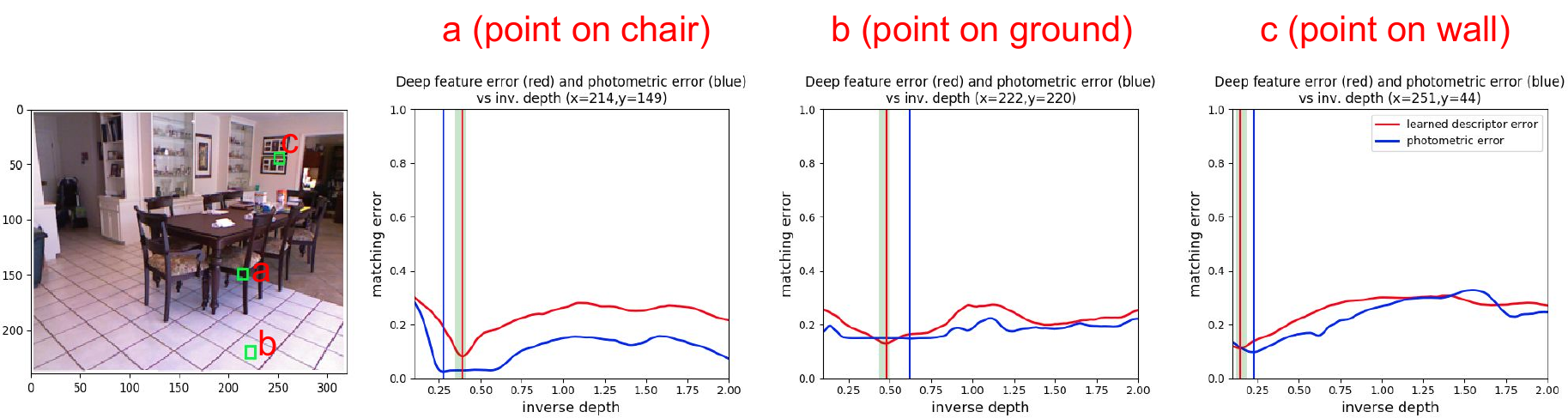}
\caption{Cost volume state (matching error vs inverse depth) for RGB features (blue) and learned deep features (red), after accumulating 30 frames, for 3 points in the keyframe.}
\label{fig:CostVolume}
\end{figure}

\begin{figure}[!t]
\centering
\includegraphics[width=\textwidth]{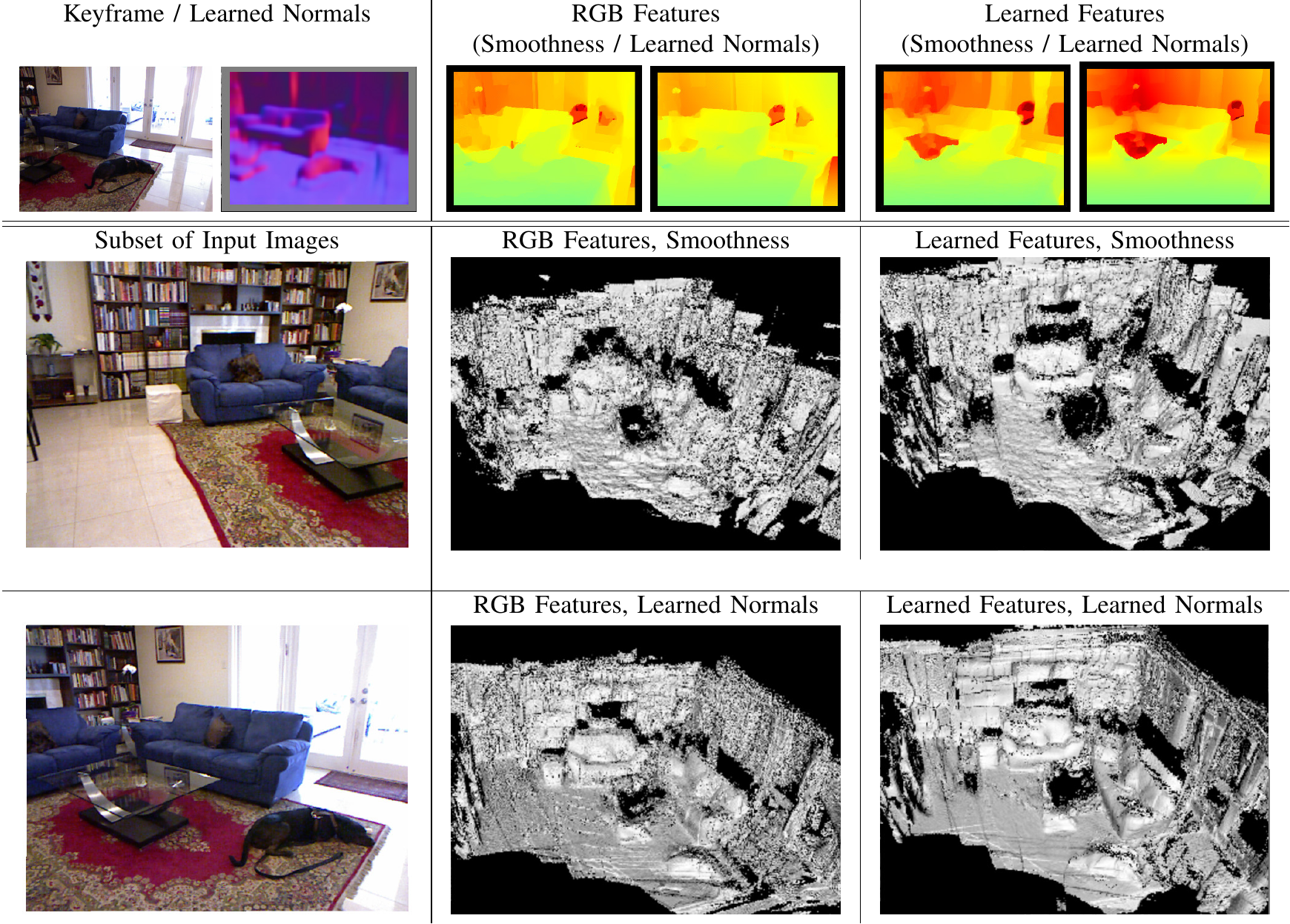}
\caption{A snapshot of our real-time live reconstruction results on the `living\_room\_0075' test sequence from NYUv2 raw dataset, as well as those of our baselines, using live monocular ORB-SLAM tracking.}
\label{fig:intro2}
\end{figure}

Since our network regresses multi-scale features, similar to \cite{xie2015holistically}, we explicitly perform deep supervision by minimizing both cross-entropy and regression losses with respect to the features at each scale independently (L1-L5), as well as the aggregated feature (L in Fig. \ref{fig:NetArch}).
The camera intrinsics are scaled to suit the output feature resolution at each block, and we obtain groundtruth depth maps at each lower resolution scale through nearest neighbour sampling of the full resolution groundtruth depth map. 
In comparison, similar works to ours like \cite{SchmidtVD,UCN16} have relied upon a single contrastive loss function at the end of the network. In addition to providing the flexibility of discarding blocks of the network in trade for computation speed, this deep supervision acts as a good regularizer for the network resulting in fast convergence during training. Our network is trained from scratch with Xavier initialization using the Caffe framework \cite{Caffe_2014}.

\subsection{Keyframe Reconstruction Using Deep Features}
We can use our learned features in the mapping frameworks of \cite{newcombe2011dtam,DenseSLAMNormals} by simply replacing the RGB cost-volume with the learned deep-features based cost-volume.
Our inference loss is:
\begin{equation}
\label{energy_total}
E(\boldsymbol{\rho}) = \operatornamewithlimits{\sum}_{p\in\mathcal{P}}
\frac{1}{\lambda}E_{\phi}(\rho_p) +  E_{reg}(\rho_p),
\end{equation}
where $E_{\phi}$ is the data-term defined as:
\begin{equation}
\label{eq:energy_data}
E_{\phi}(\rho_p)=\frac{1}{N} \operatornamewithlimits{\sum}_{n=1}^N ||\mathbf{f}_{ref}(\mathbf{u}_p) - \mathbf{f}_n(\pi(T_{nr}\pi^{-1}(\mathbf{u}_p,\rho_p)))||_1
\end{equation}
which computes the descriptor matching error for a pixel in $\mathbf{I}_{ref}$ accumulated over $N$ overlapping frames. Note here that the more robust L1 norm is used rather than square of L2 norm that was used during training. $E_{reg}$ is the regularization energy term. We experiment with the smoothness \cite{newcombe2011dtam} and normal-based \cite{DenseSLAMNormals} regularizer  as two variations. The RGB features replacing $\mathbf{f}$ in \eqref{eq:energy_data} forms \cite{newcombe2011dtam} and \cite{DenseSLAMNormals} as our baselines. $\lambda$ in each case controls the regularization strength.

\section{Experiments}
\label{sec:eval_all}
We extensively evaluate the matching performance and desirable properties of our learned deep features against several baselines on a large subset of the NYUv2 dataset \cite{SilbermanECCV12} and follow that up to further explore the generalization capability of the features quantitatively on TUM \cite{sturm12iros} and ICL-NUIM \cite{handa:etal:ICRA2014} datasets and qualitatively on KITTI \cite{KITTI} dataset. Note that our features used in all experiments are trained on the raw NYUv2 dataset, excluding all test scenes. 

\noindent \textbf{Quantitative and Qualitative Results on NYUv2:} We follow the same experimental set-up and the train/test split of NYUv2 dataset as that of \cite{DenseSLAMNormals}. Camera motion (in metric units) for all the NYUv2 train and test sequences were precomputed using \cite{ORBSLAM2} to isolate the mapping process. We use our own implementation of \cite{newcombe2011dtam} and \cite{DenseSLAMNormals} as two very strong mapping baselines for this work.
Small sub-sequences of 61 frames (30 past and 30 future frames) were used to reconstruct all the keyframes. We show the improvement in depth estimation by replacing the RGB features by our learned features for cost-volume creation in both the baselines mentioned above and compare the results to quantify the improvements. We evaluate the accuracy of the depth maps on the standard evaluation criteria used on the NYUv2 dataset by \cite{eigen2015predicting,DenseSLAMNormals}. 

To find the optimal hyper-parameter $\lambda$ for each of the 4 reconstruction approaches, we brute-force search in a sensible range. Fig. \ref{fig:VaryingLambdaNYU} visualizes the sensitivity of all the reconstruction approaches to the choice of regularization strength $\lambda$. After scaling $\lambda$ for each approach to within a constant normalization factor, it can be seen that all the approaches degrade gracefully when we deviate from the optimal choice of $\lambda$ while the learned features make this choice less critical.

In Table \ref{table:Quantitative} we report the best reconstruction accuracies obtained on NYUv2 dataset with and without using our learned features. Using our learned features to create the cost-volume gives a significant improvement to the performance (on all the evaluation measures) compared to RGB cost-volume based reconstruction. In particular the percentage of accurately reconstructed points ($\delta<1.25$) improve from 83.4\% to 93.6\% with smoothness regularization on the NYU dataset. A similar improvement is observed when learned normals are used as a prior to regularize the depth maps.

\begin{table}[!t]
\centering
\begin{tabular}{c|cccc|ccc|}
\cline{2-8}
                      & \multicolumn{4}{c}{Error (lower is better)} & \multicolumn{3}{|c|}{Accuracy (higher is better)} \\ \cline{2-8} \hline
NYU-D V2 Test Set & rms (m)  & log & abs.rel  & sq.rel & $\delta < 1.25$ & $\delta < 1.25^2$ & $\delta < 1.25^3$ \\\hline

CNN Depth\cite{eigen2015predicting}             & 0.637 & 0.226 & 0.163 & 0.135 & 0.738 & 0.937 
					  & 0.982\\\hline
P.E. + Smoothness 	  & 0.522 & 0.206 & 0.123 & 0.111 & 0.834 & 0.949 
                      & 0.979\\\hline
P.E. + L. Normals        & 0.449 & 0.174 & 0.086 & 0.076 & 0.893 & 0.964 
					  & 0.985\\\hline            
L.F.E. + Smoothness   & 0.372 & 0.143 & 0.067 & 0.054 & 0.936 & 0.978 & 0.990 \\\hline  
L.F.E. + L. Normals   & \textbf{0.356} & \textbf{0.135} & \textbf{0.058} & 	 \textbf{0.049} & \textbf{0.948} & \textbf{0.981} &  \textbf{0.991} \\\hline \hline
TUM `fr2\_desk' & rms (m)  & log & abs.rel  & sq.rel & $\delta < 1.25$ & $\delta < 1.25^2$ & $\delta < 1.25^3$ \\ \hline

CNN Depth\cite{eigen2015predicting}             & 1.141  &  0.368  & 0.227 & 0.261 & 0.543  &  0.820  & 0.923\\\hline 
P.E. + Smoothness     & 0.563 	 & 	 0.215 	 & 	0.106 	 & 	 0.091 	 & 	 0.854 	 & 	 0.924 	 & 	 0.973 \\\hline 
P.E. + Normals  &  0.558 	 & 	 0.215 	 & 	  0.103 	 & 	 0.089 	 & 	 0.863 	 & 	 0.922 	 & 	 0.969\\\hline
L.F.E. + Smoothness   &  0.453 & 0.178 & \textbf{0.079} & 0.061 & 0.909 & 0.949 	 & 	\textbf{0.981}	 \\\hline
L.F.E. + L. Normals   & \textbf{0.450} & \textbf{0.176} & \textbf{0.079} & 	\textbf{0.060} 	& \textbf{0.910} & \textbf{0.950} & \textbf{0.981} \\\hline \hline 
ICL-NUIM `lrkt0' & rms (m)  & log & abs.rel  & sq.rel & $\delta < 1.25$ & $\delta < 1.25^2$ & $\delta < 1.25^3$ \\ \hline

CNN Depth\cite{eigen2015predicting} & 0.829  &  0.426  &  0.295  & 0.261 	 &  0.472  & 0.781  & 0.905\\  \cline{1-8}
P.E. + Smoothness  & 0.287 	 & 	 0.118 	 & 	 0.062 	 & 	 0.041 	 & 	 0.943 	 & 	 0.988 	 & 	 0.997   \\  \cline{1-8}
P.E. + Normals &  0.214 	 & 	 0.094  & 	 0.047 	 & 	 \textbf{0.022} 	 & 	 0.963 	 & 	 0.993 	 & 	 \textbf{0.998} \\\hline
L.F.E. + Smoothness  & 0.285 & 0.115 & 0.056 & 0.043 & 0.952 & 0.987 & 0.996 \\\hline
L.F.E. + L. Normals  & \textbf{0.213}  & \textbf{0.090} & \textbf{0.043} & \textbf{0.022} & \textbf{0.972} 	& \textbf{0.994} & \textbf{0.998} \\\hline 
\end{tabular}
\caption[Quantitative results on indoor test sequences.]{Quantitative results on indoor test sequences. P.E. = Photometric Error. L.F.E. = Learned Feature Error. Depth and surface normal maps are predicted using our Caffe implementation of the neural network in \cite{eigen2015predicting}. The average errors and accuracy are for keyframe reconstructions against Kinect depth maps (where valid depths are available). The results here are those for the optimal $\lambda$ value of each reconstruction approach.}
\label{table:Quantitative}
\end{table}

Fig. \ref{fig:intro2} shows a visual comparison of the reconstructions obtained using our proposed deep features against those obtained using RGB features, with either learned normals or smoothness prior. The reconstructions were generated in near real-time in parallel with ORB-SLAM monocular visual tracking \cite{ORBSLAM_2015}. We fuse the depth maps of the keyframes obtained using each of the approaches using InfiniTAM \cite{InfiniTAM} as part of our framework for better visualization and the results are shown and analysed in the figures. A significant advantage of the proposed deep features can be easily seen in these results visually. 

\begin{figure}[!t]
\centering
\includegraphics[width=\textwidth,clip]{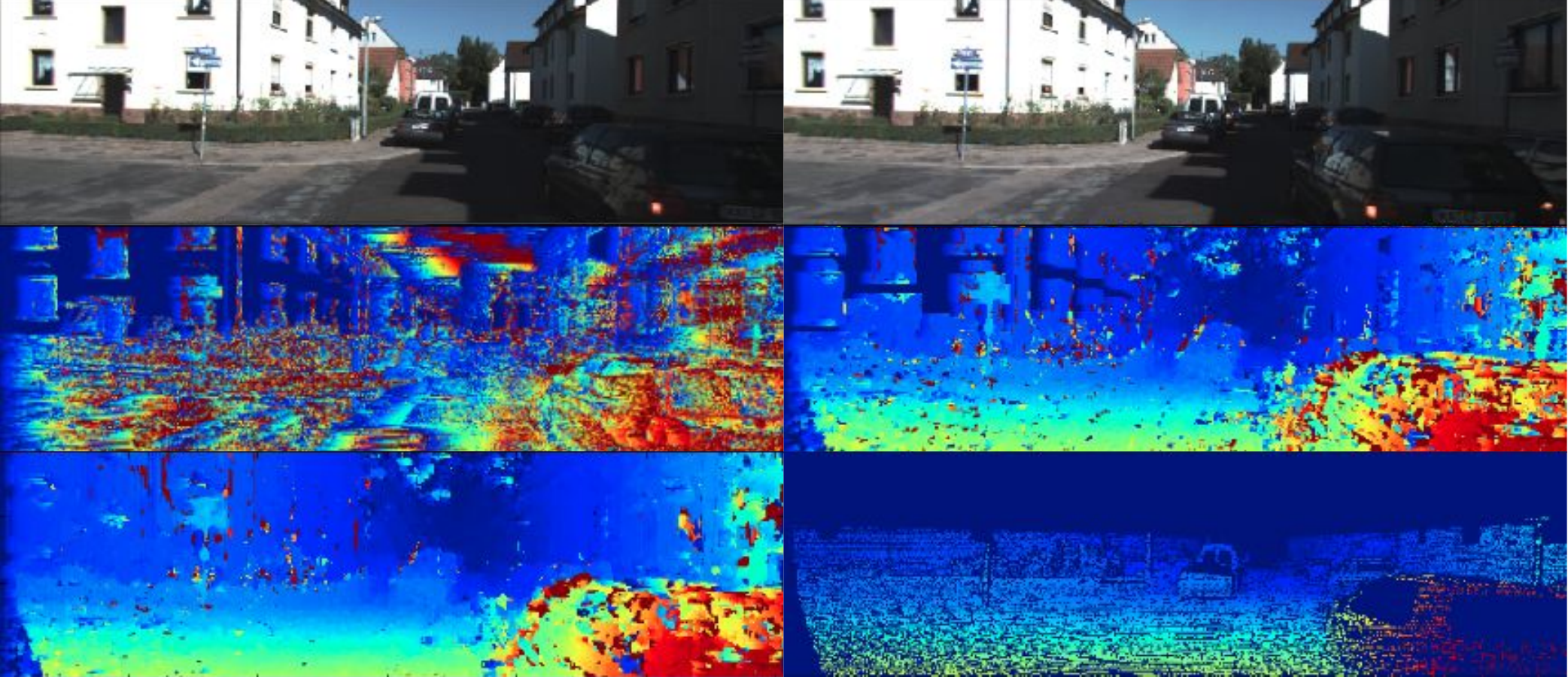}
\caption{Generalizability of learned features trained on NYU dataset to a completely different type of scene on the KITTI stereo dataset without any finetuning. Note that this is just the pure matching result with no regularization. From Left-Right, Top-Bottom: Reference Image, Live Image, RGB-features based disparity map, Block B1 features-based disparity map, Full network features-based disparity map, Groundtruth.}
\label{fig:kitti}
\end{figure}

\noindent \textbf{Generalization Capabilities Across Different Camera Models:} Furthermore we quantify the importance of the learned context aware deep features over RGB matching on two more indoor datasets, TUM dataset `fr2$\_$desk' and ICL-NUIM synthetic dataset `lr  kt0', both of which contain images that possess different properties (e.g. camera intrinsics) from that of NYUv2 raw.
As seen in the bottom part of Table \ref{table:Quantitative}, the deep features consistently improve the depth estimation accuracy on these datasets without need for fine-tuning. The generalization capability of our network can be clearly observed in the example shown in Fig. \ref{fig:kitti} where KITTI stereo images are matched using our features learned on the NYU dataset. In this example the camera model, lighting conditions, objects and textures present in the scene, and camera motion significantly vary from the indoor setup. Also notice in the figure how explicitly combining course features with the fine features provides the necessary context to match large textureless regions while maintaining precision.

\noindent \textbf{Comparison With Other Feature Matching Techniques:}
We compare our learned features against some baselines which are also suited for the dense geometric matching task with their own unique strengths and weaknesses. Apart from raw RGB features as commonly used in dense SLAM systems like DTAM, we chose to use as baselines: 7x7 patches that are affine warped before matching (similar to what is used in DSO \cite{DSO}), ResNet-50 Conv1 features \cite{he2016deep} pre-trained on ImageNet (with a receptive field of 7x7), DenseSIFT (used for dense optic flow in \cite{liu2016sift}), and features from Block B1 of our network (with a receptive field of 7x7). Results of keyframe reconstructions (based purely on matching with 30 frames) for some examples in and out of NYU dataset are shown in Fig. \ref{fig:FeatComp1_1} and the rest in supplementary material. The supplementary material also contains 2-frame matching results.

From the results it can be seen that our learned features (particularly the full network version) consistently produce reconstructions close to groundtruth, and are superior to the baselines in terms of both matching accuracy and precision with minimal piecewise constant depth artifacts that occur in some of the baselines when overlapping images have severe appearance distortion. Although in some examples the number of accurate matches is higher in some of the baselines when matching with two frames, the number of accurate matches when matching based on 30 images is consistently higher using our full learned features. This indicates that the matching performance of the baselines is biased towards particular types of appearance changes present in the overlapping images which is an undesirable property. Please refer supplementary material for the complete set of visual and quantitative comparisons.

\noindent \textbf{A Detailed Analysis of Learned Deep Features:} A deeper analysis of the cost volumes generated using the proposed learned features (Fig. \ref{fig:CostVolume}) verses those generated using using RGB features highlight three main advantages of using our learned deep features. Particularly, our learned deep features-based cost volume has (1) sharp (non-flat) minima which leads to unambiguous unique matching even in textureless areas without having to rely on priors, (2) small number of local minima leading to a large basin of convergence and favoring gradient based optimization methods which heavily rely on priors and initialization, and most importantly, (3) the global minima corresponds to the correct inverse depth solution in majority of the cases, even with only a few number of overlapping frames (as shown in more detail in the supplementary material). 

\section{Conclusion} 
\label{sec:conclusion}

In this work we have presented a novel efficient Convolutional Neural Network architecture and a method for learning context-aware deep features which are ``good features to match" in the context of dense mapping. We presented an extensive visual analysis which highlight some of the desirable properties of these deep features (invariance to illumination and viewpoint changes and ability to uniquely match thanks to the large receptive field). 
With the help of extensive quantitative analysis on three different datasets it was shown that the learned features generalize well across data captured with different cameras to give state-of-the-art reconstructions of indoor scenes in real-time. Initial experiments show promising stereo matching results even in substantially different outdoor scenes of the KITTI dataset where not only the camera but the brightness as well as the scene contents change substantially. We would like to further test the generalization capability of our learned features to different domains (outdoor scenes), and other vision tasks (image classification, retrieval, etc.) and explore if the learned features can be used for accurate camera tracking in a direct image registration framework.


\end{document}



\title{Learning Deeply Supervised Good Features to Match for Dense Monocular Reconstruction (Supplementary Material)} 
\titlerunning{Learning Deeply Supervised Good Features to Match (Supplementary)} 


\author{Chamara Saroj Weerasekera \inst{1,2} \and
Ravi Garg \inst{1,2} \and
Yasir Latif \inst{1,2} \and Ian Reid \inst{1,2}}
%

\authorrunning{C. S. Weerasekera et al.} 


\institute{University of Adelaide, Australia 
\and
ARC Centre of Excellence for Robotic Vision
\email{firstname.lastname@adelaide.edu.au}}

\maketitle

\noindent \textbf{Comparison With Other Feature Matching Techniques Contd.:}

\begin{figure}
\centering
\includegraphics[width=0.95\textwidth]{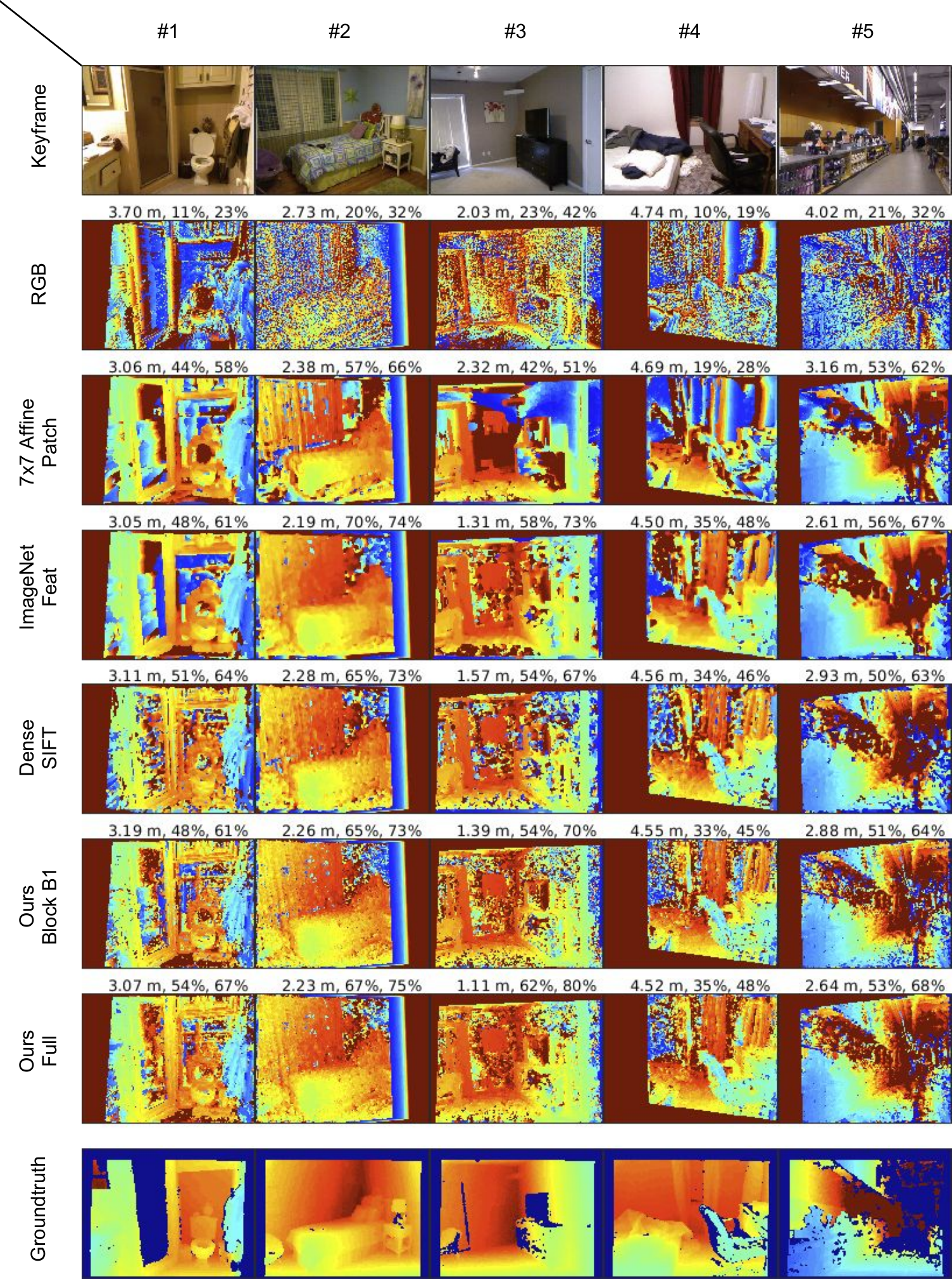}
\caption{Example set 1 comparing keyframe reconstructions (based purely on matching against 1 frame) using our learned features against those of baselines. The three numbers above each result denote RMSE (m), high precision accuracy ($\delta < 1.1$), and medium precision accuracy ($\delta<1.25$).}
\label{fig:FeatComp1_1}
\end{figure}

\begin{figure}
\centering
\includegraphics[width=0.95\textwidth]{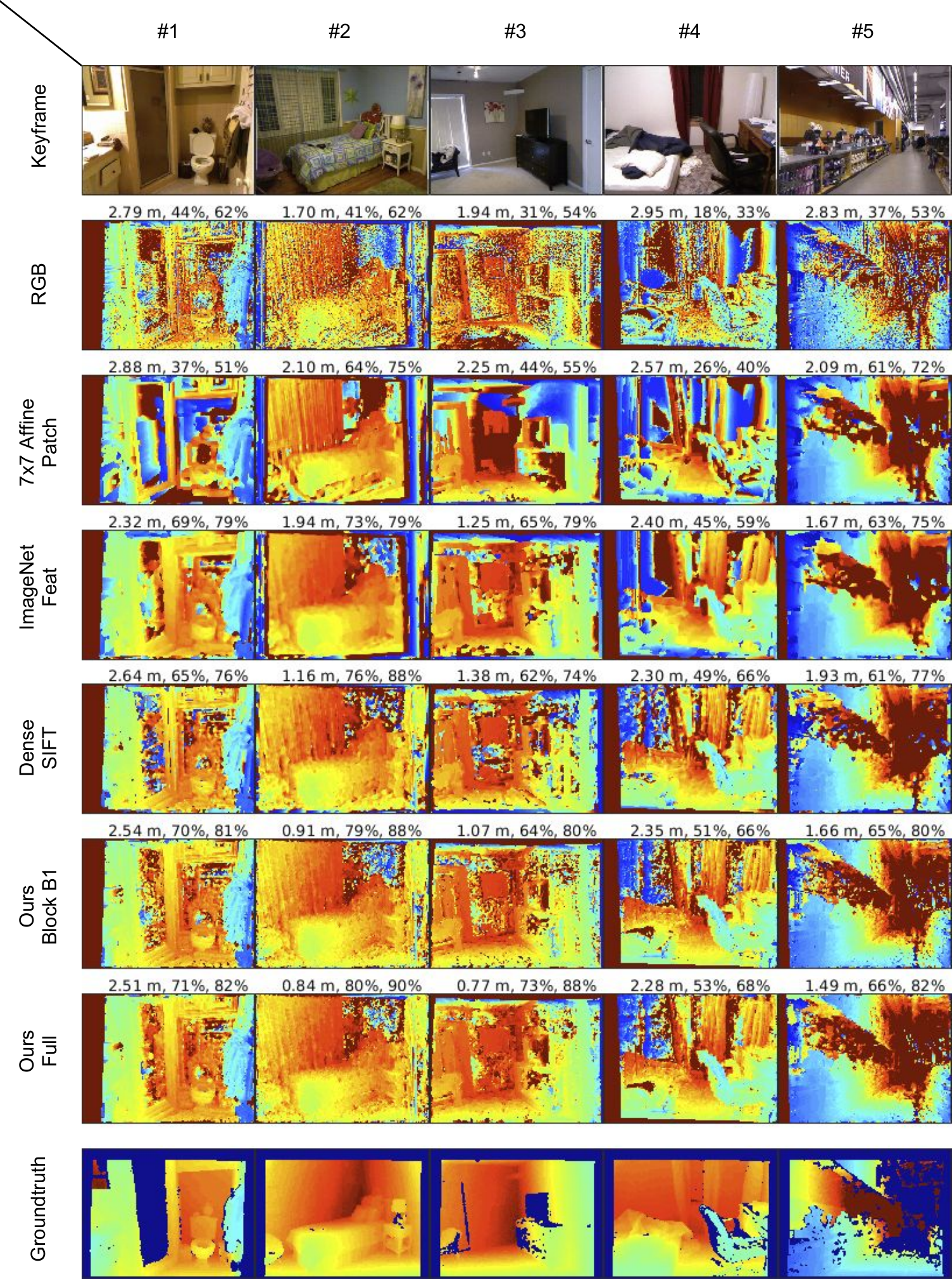}
\caption{Example set 1 comparing keyframe reconstructions (based purely on matching against 30 frames) using our learned features against those of baselines. The three numbers above each result denote RMSE (m), high precision accuracy ($\delta < 1.1$), and medium precision accuracy ($\delta<1.25$).}
\label{fig:FeatComp1_30}
\end{figure}

\begin{figure}
\centering
\includegraphics[width=0.95\textwidth]{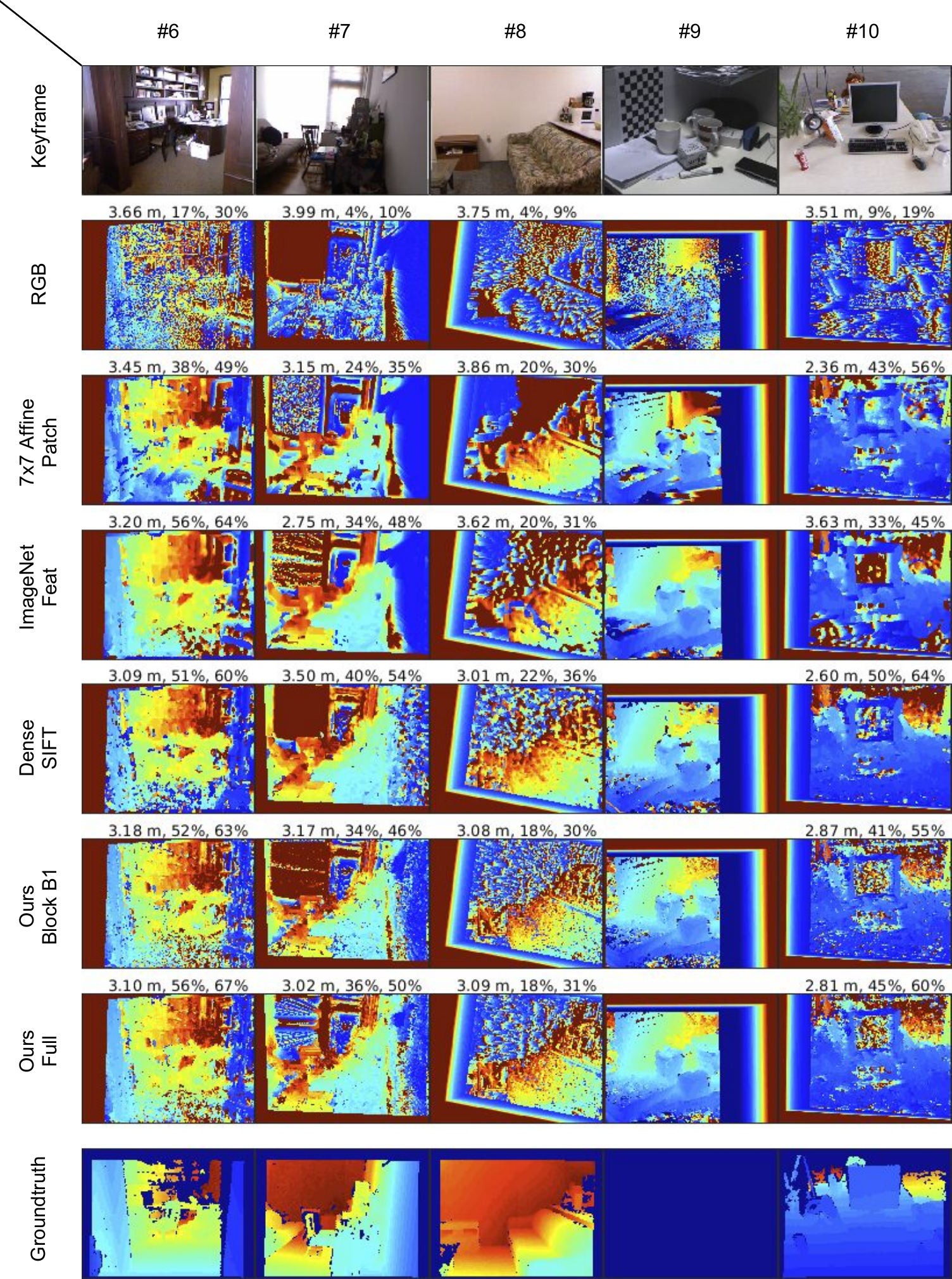}
\caption{Example set 2 comparing keyframe reconstructions (based purely on matching against 1 frame) using our learned features against those of baselines. The three numbers above each result denote RMSE (m), high precision accuracy ($\delta < 1.1$), and medium precision accuracy ($\delta<1.25$).}
\label{fig:FeatComp2_1}
\end{figure}

\begin{figure}
\centering
\includegraphics[width=0.95\textwidth]{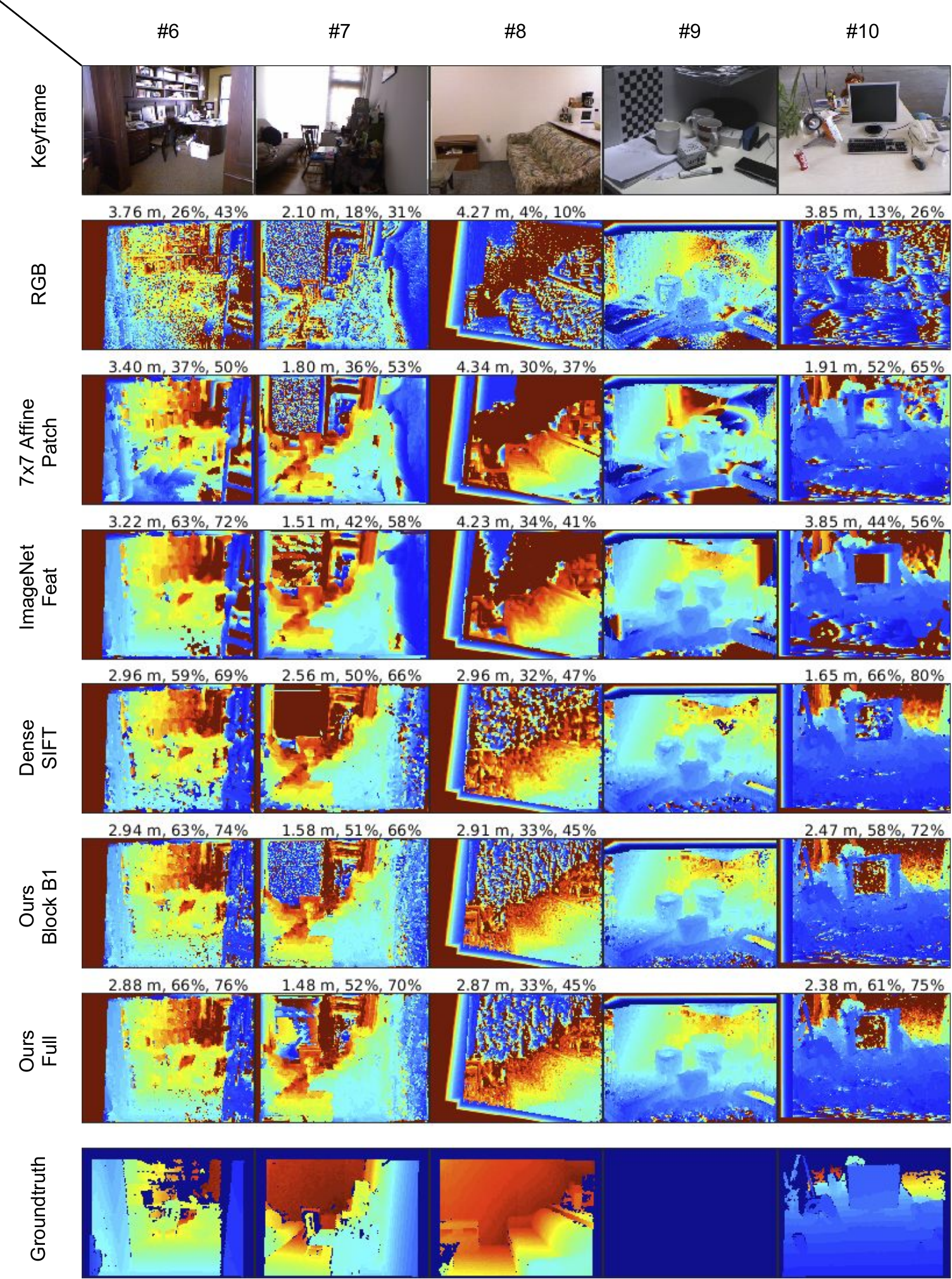}
\caption{Example set 2 comparing keyframe reconstructions (based purely on matching against 30 frames) using our learned features against those of baselines. The three numbers above each result denote RMSE (m), high precision accuracy ($\delta < 1.1$), and medium precision accuracy ($\delta<1.25$).}
\label{fig:FeatComp2_30}
\end{figure}

\begin{figure}
\centering
\includegraphics[width=0.6\textwidth]{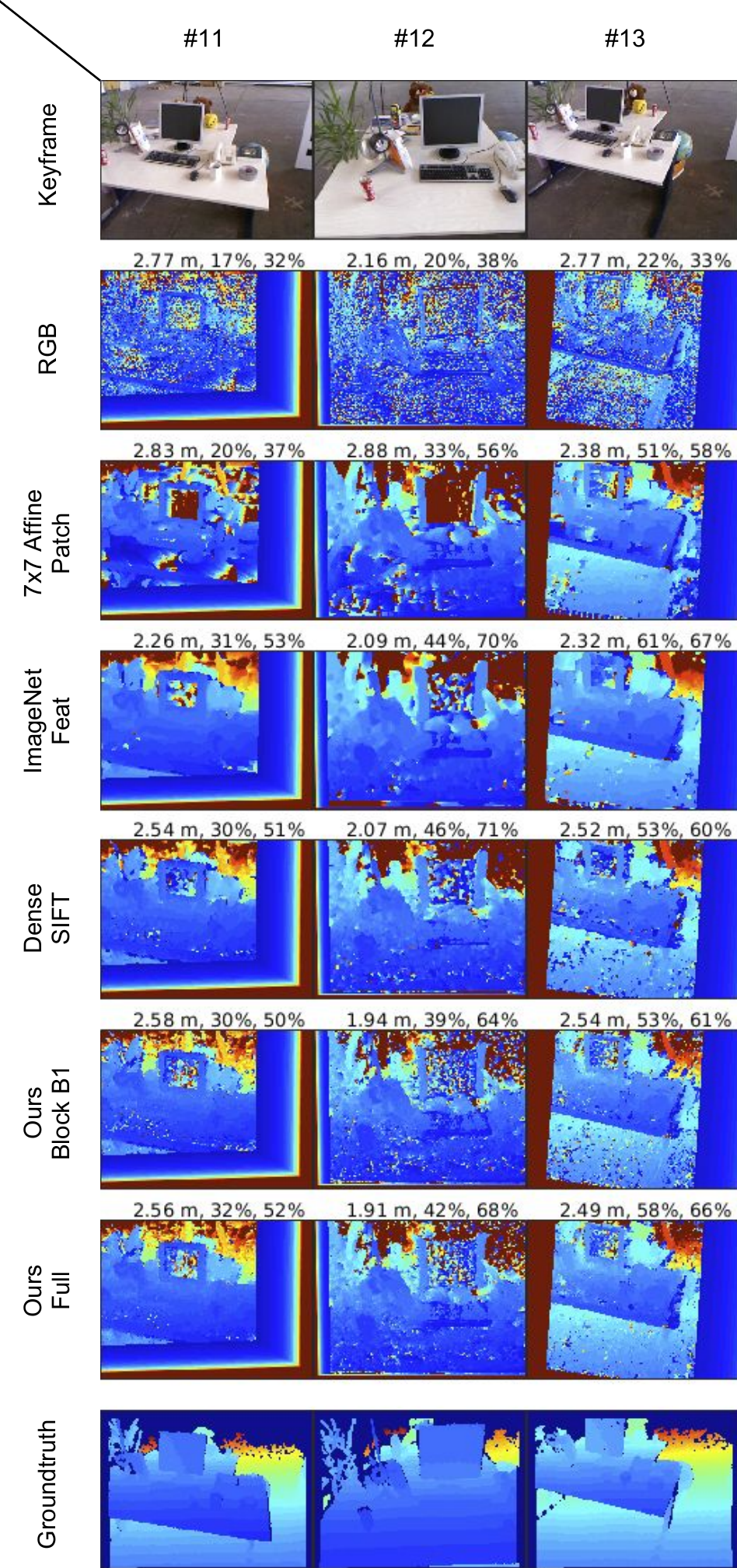}
\caption{Example set 3 comparing keyframe reconstructions (based purely on matching against 1 frame) using our learned features against those of baselines. The three numbers above each result denote RMSE (m), high precision accuracy ($\delta < 1.1$), and medium precision accuracy ($\delta<1.25$).}
\label{fig:FeatComp3_1}
\end{figure}

\begin{figure}
\centering
\includegraphics[width=0.6\textwidth]{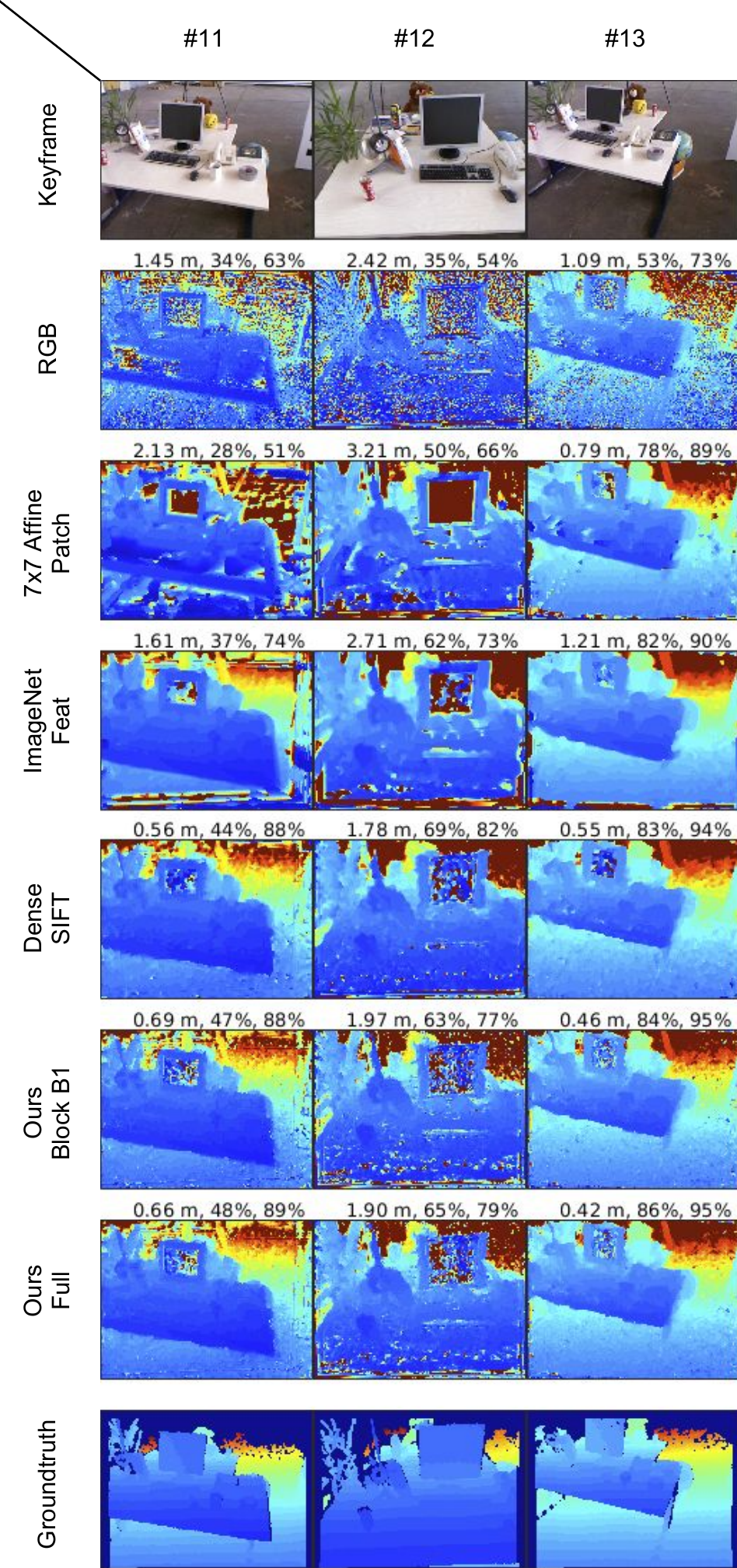}
\caption{Example set 3 comparing keyframe reconstructions (based purely on matching against 30 frames) using our learned features against those of baselines. The three numbers above each result denote RMSE (m), high precision accuracy ($\delta < 1.1$), and medium precision accuracy ($\delta<1.25$).}
\label{fig:FeatComp3_30}
\end{figure}

Figures \ref{fig:FeatComp1_1}, \ref{fig:FeatComp1_30}, \ref{fig:FeatComp2_1}, \ref{fig:FeatComp2_30}, \ref{fig:FeatComp3_1}, and \ref{fig:FeatComp3_30} compare our learned features against our baselines: Raw RGB features, 7x7 patches that are affine warped before matching (similar to what is used in DSO \cite{DSO}), ResNet-50 Conv1 features \cite{he2016deep} pre-trained on ImageNet (with a receptive field of 7x7), DenseSIFT (used for dense optic flow in \cite{liu2016sift}), and features from Block B1 of our network (with a receptive field of 7x7). 

The only example where a baseline (DenseSIFT) outperforms our learned features with 30 frames is in the TUM `fr2\_xyz' sequence (Example $\#10$ and $\#12$ in Figures \ref{fig:FeatComp2_30} and \ref{fig:FeatComp3_30}) where no camera rotation is observed. We believe this compromise is a consequence of enforcing our learned features to match across images acquired with a camera undergoing \emph{general} motion during training, where more severe appearance distortion can occur unlike in the stereo or xyz motion case. On the other hand, if we allowed the feature network to learn features specifically for the stereo or xyz motion case, it may have outperformed DenseSIFT in such cases. Nevertheless even in the xyz motion case the performance of our learned features is not too far behind DenseSIFT and often greatly outperform 7x7 affine patch features and ResNet-50 ImageNet Conv1 features in matching performance.

\noindent \textbf{A Detailed Analysis of Learned Deep Features Contd.:} 

\begin{figure}
\centering
\includegraphics[width=0.8\textwidth]{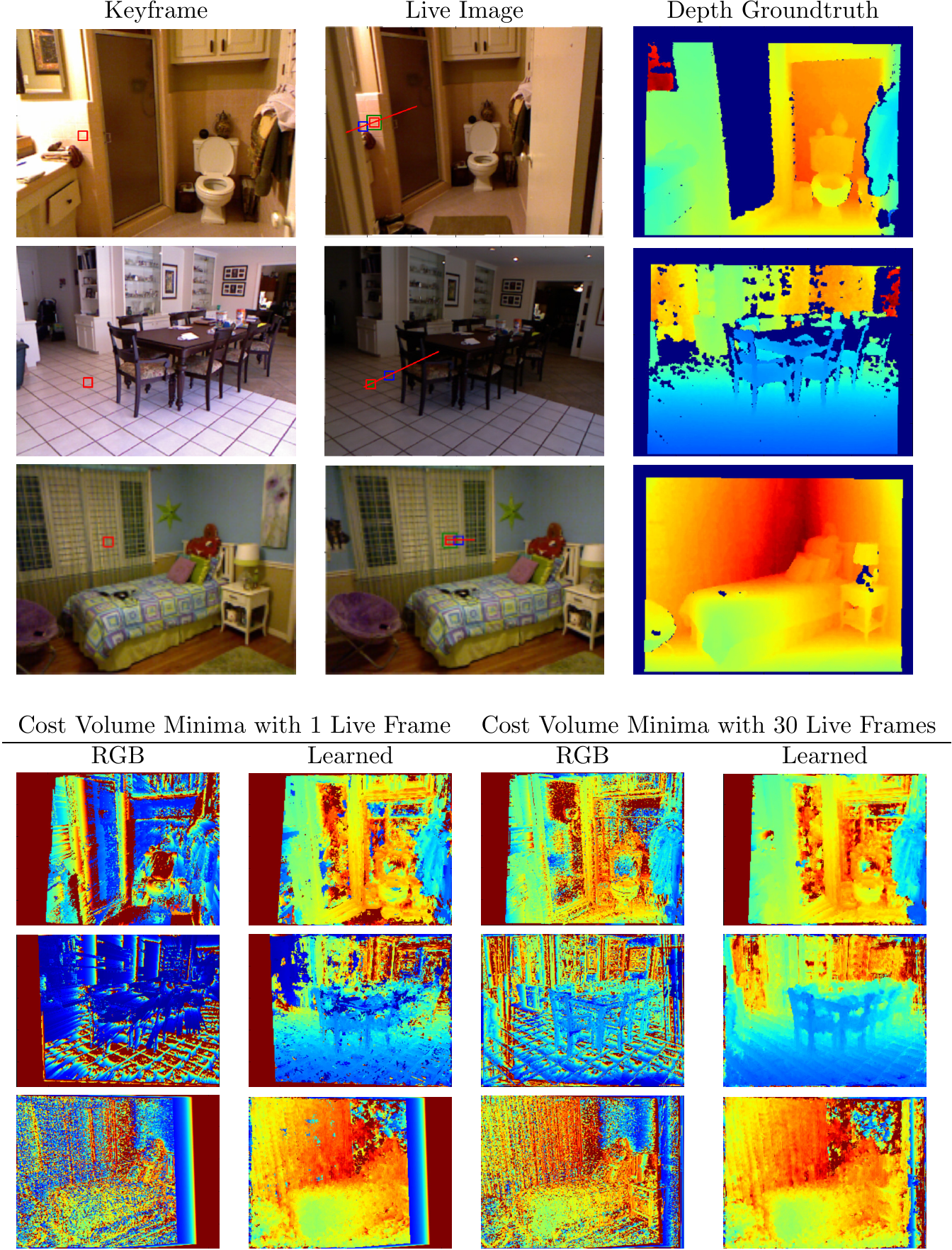}
\caption{Figure compares RGB and learned features for keyframe reconstruction purely based on feature matching for: a bathroom, a dining room, and a bedroom sequence from the NYUv2 dataset. \textbf{Top:}
The red square in the first image in each row is a point of interest, which has to be mapped to the corresponding image in the second column on the epipolar line (denoted in red). The red square in the second image correspond to the predicted match using deep features which aligns correctly with the true match (green square) while the blue square represents the best RGB match. The last column is the ground truth depth map for each case.
\textbf{Bottom:} The left column shows the depth maps obtained by matching the two frames shown in the top using RGB and learned features respectively. It is evident that, the learned features are more reliable for matching two frames and successfully match far more points correctly despite absence of local texture. The right column shows the depth maps obtained by minimizing RGB-based and learned features-based cost volumes respectively given 30 live frames. Matching evidence gets accumulated over frames for both cases but the cost volume constructed using the learned features gives considerably better results.}
\label{fig:CostVmin}
\end{figure}

Fig. \ref{fig:CostVmin} visualizes the depth map solutions inferred from the cost volume generated using RGB features with that generated using our learned features in some challenging scenarios involving large baseline matching along epipolar lines. Even when only two frames are used for matching, the learned features --- consisting of contextual information from a large receptive field --- can be successful matched, while RGB features often struggle to establish a good match as expected. As the number of frames available for matching grows, very realistic and smooth depth maps can be inferred directly from our learned features-based cost volume without the application of any prior.

\begin{figure}
\centering
\includegraphics[width=0.9\textwidth]{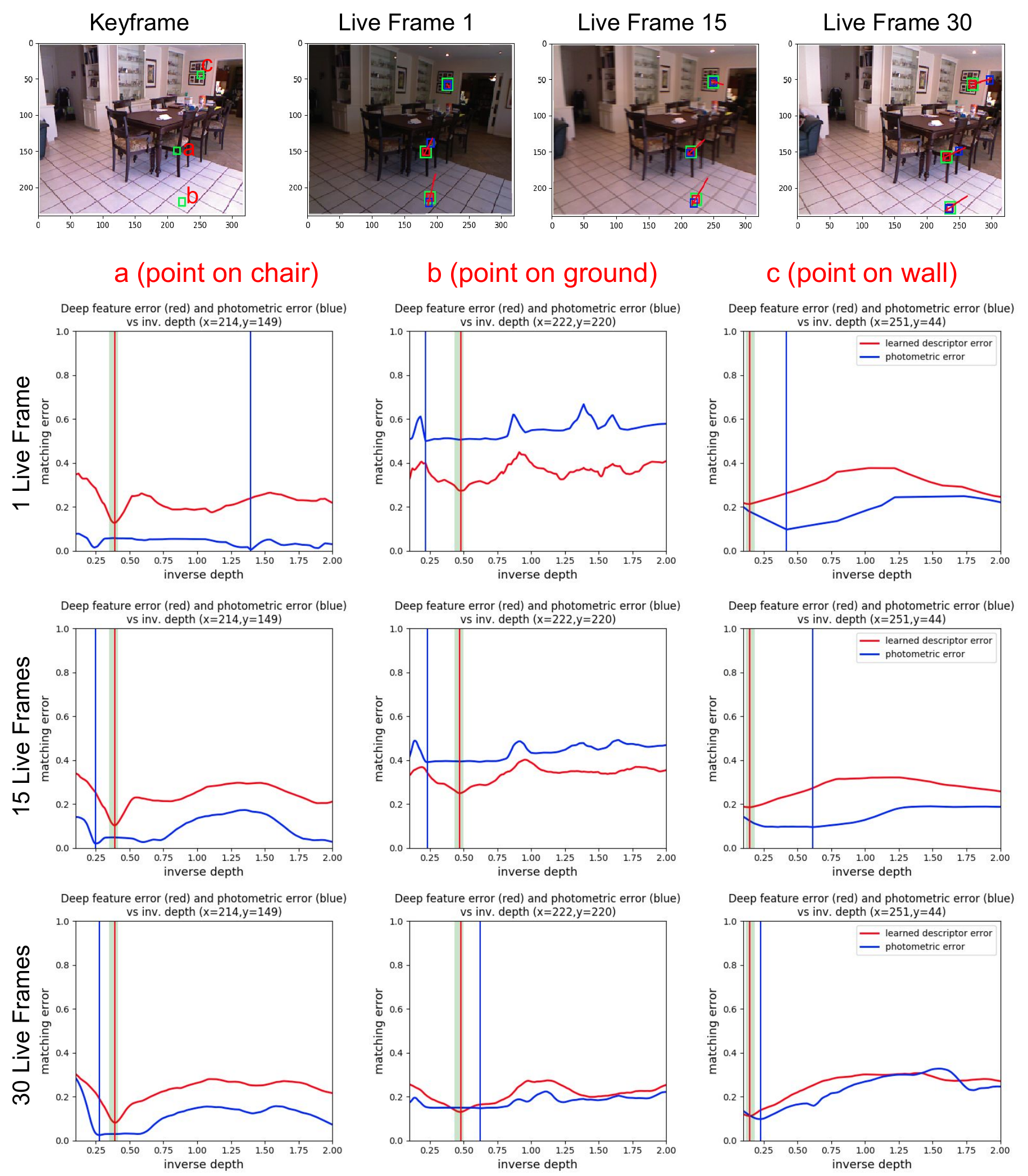}
\caption{Figure shows the cost-volume as it evolves after accumulating evidences from multiple frames for 3 different points, using learned features (red) and rgb features (blue). The red and blue vertical lines in the plots indicate the location of the minimum using learned features and rgb features respectively, and the green vertical line indicates the location of the groundtruth minimum. From top to bottom, the plots correspond to cost volumes accumulated over 2 frames, 15 frames and 30 frames respectively. In the top images, the green squares show the location of the 3 points of interest in the reference image (left), and the corresponding groundtruth locations in 3 of the live images that were used to generate the cost volume. The red and blue squares show the matched locations of the same 3 points after searching along the corresponding epipolar lines in the live frames using learned features and rgb features respectively.}
\label{fig:costVplot}
\end{figure}

Fig. \ref{fig:costVplot} highlights in more detail the main advantages of using our learned features for matching. Particularly, our learned deep features-based cost volume have (i) sharp (non-flat) minima which leads to unambiguous unique matching even in textureless areas, (ii) small number of local minima leading to a large basin of convergence and favoring gradient based optimization methods which heavily rely on priors and initialization, and most importantly, (iii) the global minima corresponds to the correct inverse depth solution in majority of the cases, even with only a few number of overlapping frames.


